\documentclass[11pt]{article}

\PassOptionsToPackage{dvipsnames,table}{xcolor}
\usepackage[preprint]{acl}
\usepackage{booktabs} % 提供顶级横线命令
\usepackage{multirow} % 提供跨行支持
\usepackage{makecell} % 提供单元格内换行
\usepackage{adjustbox} % For adjustbox environment
\usepackage[most]{tcolorbox}
\usepackage{booktabs}
\usepackage{amsmath}
\tcbuselibrary{skins}
 \usepackage{multicol}
\usepackage{tabularx}
\usepackage{booktabs}
\usepackage{multirow}
\usepackage{enumitem} 
\usepackage{multirow}
\newtcolorbox{taskbox}[2][]{
	enhanced, breakable,
	colframe=blue3!40,
	colback=blue5!5,
	arc=1mm,
	outer arc=1mm,
	fontupper=\small,
	fontlower=\small,
	coltitle=blue1,
	fonttitle=\bfseries,
	boxsep=1mm,
	left=0mm,
	right=0mm,
	top=0mm,
	bottom=0mm,
	before={\noindent},
	segmentation style={solid, blue3},
	title=#2,
	#1
}
\newtcolorbox{mybox}[2][]{
	width=\textwidth,
	colback = gray!8, 
	colframe = gray!8, 
	boxsep=0pt,left=12pt,right=12pt,top=12pt,bottom=12pt,
	fontupper=\linespread{0.9}\selectfont,
	title=#2,#1}

\newcolumntype{Y}{>{\centering\arraybackslash}X}
\newcommand{\std}[1]{{\scriptsize $\pm$ #1}}
\definecolor{googleblue}{HTML}{4285F4}
\definecolor{googlegray}{HTML}{F8F9FA}
\usepackage{tabularx}   % 自动列宽
\usepackage{booktabs}   % 专业横线
\usepackage{colortbl}   % 行颜色
\usepackage{multirow}   % 跨行支持
\usepackage{amsmath}    % 数学符号
\usepackage{times}
\usepackage{latexsym}
\usepackage{pifont}
\usepackage[T1]{fontenc}
\usepackage{amsmath}   % 提供了 \begin{equation}, \text{} 等基础数学环境
\usepackage{amssymb}   % 提供了 \mathbb{R} (实数集R), \mathcal{} (花体) 等数学符号
\usepackage{amsfonts}  % 字体支持，通常与 amssymb 配合使用
\usepackage{bm}        % 提供了 \boldsymbol{}，用于公式中的粗体希腊字母 (如 \boldsymbol{\xi})
\usepackage{booktabs}   % 顶级横线
\usepackage{multirow}   % 跨行
\usepackage{colortbl}   % 行/列背景色
\usepackage{amsmath}    % 数学符号
\usepackage{pifont}    % 提供了 \ding{182} (❶) 这类带圈数字符号

\usepackage{graphicx}  % 支持插入图片 (Figure 1, 2)

\usepackage{hyperref}  % 让 \ref{}, \cite{} 生成可点击的超链接 (通常放在最后加载)
\usepackage[utf8]{inputenc}

\usepackage{microtype}
\usepackage{inconsolata}

\usepackage{graphicx}

\def\method{\textsc{OmniFlow}} 
\title{\raisebox{-0.15em}{\includegraphics[height=1.3em]{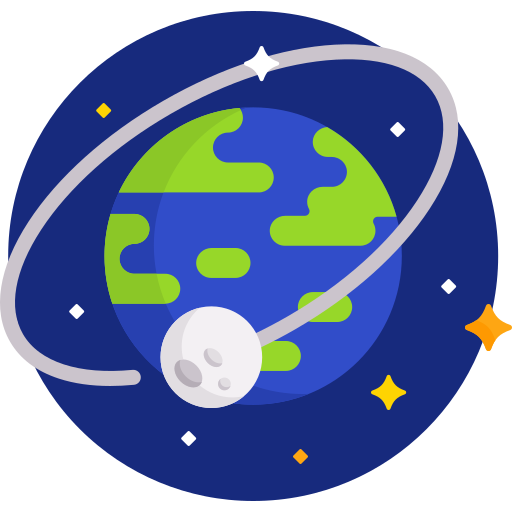}} \method{}: A Physics-Grounded Multimodal Agent for Generalized Scientific Reasoning}

\author{
  Hao Wu\textsuperscript{1,*},
  Yongheng Zhang\textsuperscript{2,*},
  Yuan Gao\textsuperscript{1,*},
  Fan Xu\textsuperscript{3},
  Fan Zhang\textsuperscript{2}
\\
  \textbf{Ruobing Xie}\textsuperscript{\textbf{2}},
  \textbf{Ruijian Gou}\textsuperscript{\textbf{4}},
  \textbf{Yuxuan Liang}\textsuperscript{\textbf{5}},
  \textbf{Xiaomeng Huang}\textsuperscript{\textbf{1}},
  \textbf{Xian Wu}\textsuperscript{\textbf{2},$\dagger$}
\\[0.35em]
  {\normalfont \textsuperscript{1}Tsinghua University \quad
  \textsuperscript{2}Tencent \quad
  \textsuperscript{3}Shenzhen Loop Area Institute}
\\
  {\normalfont \textsuperscript{4}Ocean University of China \quad
  \textsuperscript{5}HKUST (Guangzhou)}
\\[0.15em]
  {\normalfont \small \textsuperscript{*}Equal contribution \quad \textsuperscript{$\dagger$}Corresponding author}
\\[0.15em]
  {\normalfont \small \href{mailto:wuhao2022@mail.ustc.edu.cn}{Main Contact: wuhao2022@mail.ustc.edu.cn}}
}

\makeatletter
\AtBeginDocument{
\def\@maketitle{
  \vbox{
    \hsize\textwidth
    \linewidth\hsize
    \vskip 0.05in
    \centering
    {\Large\bfseries \@title \par}
    \vskip 0.12in
    {\def\and{\unskip\enspace{\rmfamily and}\enspace}%
     \def\And{\end{tabular}\hss \egroup \hskip 1in plus 2fil
              \hbox to 0pt\bgroup\hss \begin{tabular}[t]{c}\bfseries}%
     \def\AND{\end{tabular}\hss\egroup \hfil\hfil\egroup
              \vskip 0.12in
              \hbox to \linewidth\bgroup\large \hfil\hfil
              \hbox to 0pt\bgroup\hss \begin{tabular}[t]{c}\bfseries}%
     \hbox to \linewidth\bgroup\large \hfil\hfil
       \hbox to 0pt\bgroup\hss
         \outauthor
       \hss\egroup
     \hfil\hfil\egroup}
    \vskip 0.12in
  }
}
}
\makeatother

\begin{document}
\maketitle

\begin{abstract}
Large Language Models (LLMs) have demonstrated exceptional logical reasoning capabilities but frequently struggle with the continuous spatiotemporal dynamics governed by Partial Differential Equations (PDEs), often resulting in non-physical hallucinations. Existing approaches typically resort to costly, domain-specific fine-tuning, which severely limits cross-domain generalization and interpretability. To bridge this gap, we propose \method{}, a neuro-symbolic architecture designed to ground frozen multimodal LLMs in fundamental physical laws without requiring domain-specific parameter updates. \method{} introduces a novel \textit{Semantic-Symbolic Alignment} mechanism that projects high-dimensional flow tensors into topological linguistic descriptors, enabling the model to perceive physical structures rather than raw pixel values. Furthermore, we construct a Physics-Guided Chain-of-Thought (PG-CoT) workflow that orchestrates reasoning through dynamic constraint injection (e.g., mass conservation) and iterative reflexive verification. We evaluate \method{} on a comprehensive benchmark spanning microscopic turbulence, theoretical Navier-Stokes equations, and macroscopic global weather forecasting. Empirical results demonstrate that \method{} significantly outperforms traditional deep learning baselines in zero-shot generalization and few-shot adaptation tasks. Crucially, it offers transparent, physically consistent reasoning reports, marking a paradigm shift from black-box fitting to interpretable scientific reasoning. Our code is available at \url{https://github.com/Alexander-wu/OMNIFLOW}.
\end{abstract}    

\section{Introduction}
\label{sec:intro}

Large Language Models (LLMs)~\cite{chang2024survey,naveed2023comprehensive,zhao2023survey} have demonstrated exceptional symbolic reasoning, code generation, and mathematical problem-solving capabilities. However, when applied to the physical world governed by Partial Differential Equations (PDEs)~\cite{chen2001partial, evans2010partial}, particularly systems involving continuous spatiotemporal dynamics~\cite{yu2018spatio,mohan2020spatio} such as turbulence evolution~\cite{davidson2015turbulence, pope2001turbulent} or global weather forecasting~\cite{rasp2020weatherbench, gao2022earthformer, wu2025triton, gao2025oneforecast}, LLMs often exhibit significant limitations. Lacking physical grounding, existing multimodal models struggle to comprehend the topological structures within high-dimensional fluid data, frequently resulting in non-physical hallucinations that violate fundamental physical common sense.

\begin{figure}[t]
    \centering
    \includegraphics[width=0.48\textwidth]{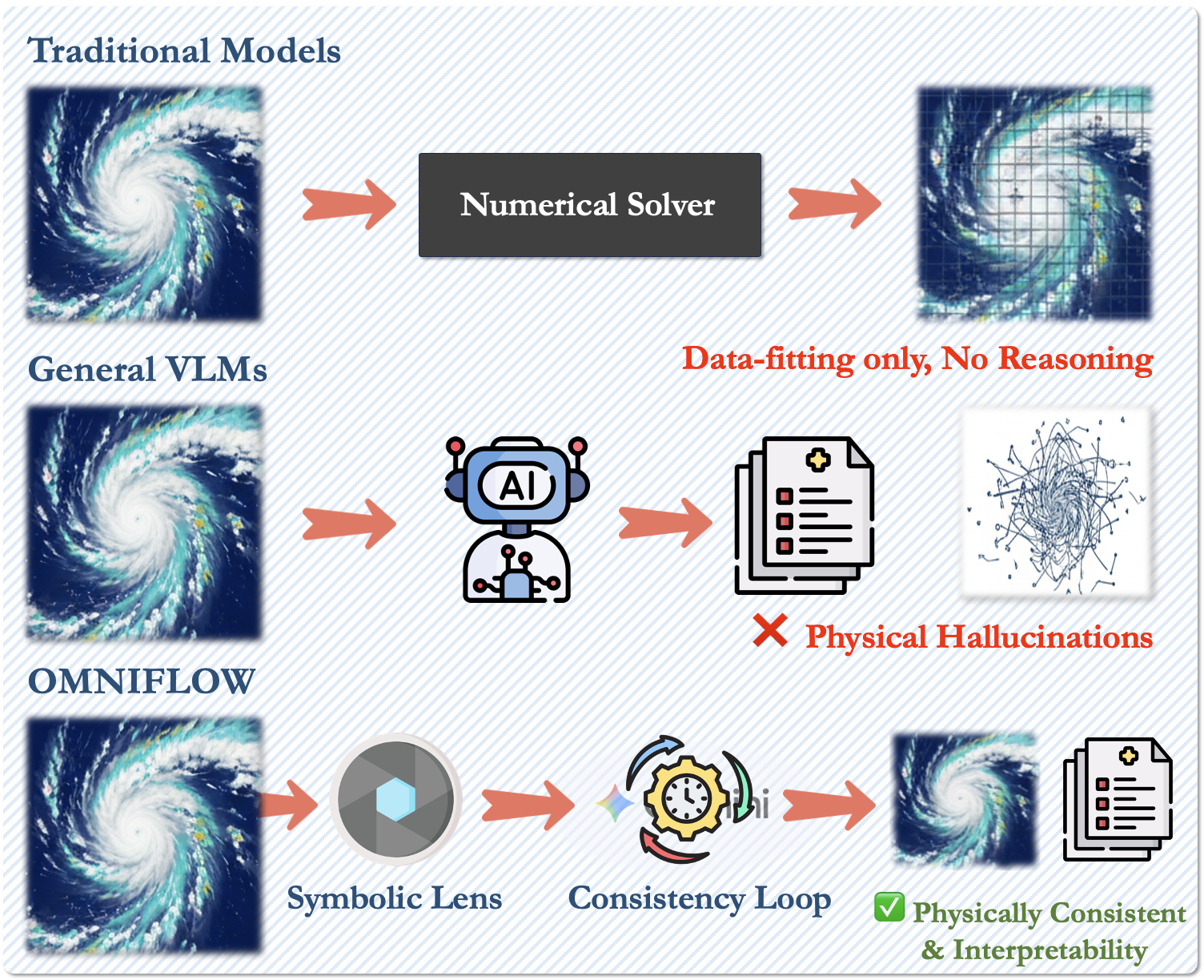} 
\caption{\textbf{Comparison of reasoning paradigms.} Traditional models (top) are non-interpretable black-boxes. General VLMs (middle) suffer from physical hallucinations. \method{} (bottom) integrates a \textit{Symbolic Lens} and \textit{Consistency Loop} to deliver grounded forecasts and expert reports, uniting numerical precision with logical reasoning.}
\label{fig:intro_comparison}
\end{figure}

Historically, two main paradigms have addressed this challenge. \ding{182}. The first involves specialized deep learning models (e.g., FNO~\cite{li2020fourier}, GraphCast~\cite{lam2023learning}), which serve as surrogates for numerical solvers. While accurate, these models function primarily as data fitters, lacking cross-domain generalization capabilities and interpretability. \ding{183}. The second paradigm involves fine-tuning LLMs with large-scale scientific data. However, fine-tuning incurs high computational costs, often leads to catastrophic forgetting of general knowledge, and still fails to guarantee strict adherence to physical conservation laws, such as mass or momentum conservation~\cite{du2024conditional}. As illustrated in Figure~\ref{fig:intro_comparison}, even advanced Vision-Language Models (VLMs) fall into this second category's trap: they interpret scientific imagery as semantic patterns rather than discrete solutions to PDEs, leading to visually plausible but physically invalid outputs.

We argue that the inability of LLMs to handle physical problems stems not from a lack of internal knowledge, but from a lack of modal alignment and logical constraint. Existing general-purpose VLMs fundamentally lack the numerical precision and physical inductive bias required for spatiotemporal fluid forecasting. Consequently, relying solely on standard VLMs leads to prognostic outputs that strictly violate conservation laws. To address this, we explore a novel paradigm: instead of modifying model parameters, we design a Cognitive Architecture to decouple physical computation from cognitive reasoning, aligning frozen LLMs with rigorous physical rules via neuro-symbolic collaboration.

To this end, we propose \method{}, a physics-grounded agent framework for generalized fluid dynamics reasoning. As shown in Figure~\ref{fig:intro_comparison}, \method{} abandons the traditional black-box prediction mode~\cite{raissi2019physics, li2020fourier, bi2023accurate, wu2024earthfarsser} in favor of a transparent reasoning workflow. First, addressing the heterogeneity of multimodal inputs~\cite{dosovitskiy2021image,liu2024visual}, we design a Visual Symbolic Projector (the \textit{Symbolic Lens} in Figure~\ref{fig:intro_comparison}). This module translates raw flow fields (e.g., satellite typhoon imagery) into semantic tokens containing vector field features and topological skeletons, achieving alignment between continuous data and discrete symbols. Second, we introduce a Physics-Guided Chain-of-Thought (PG-CoT)~\cite{wei2022chain}. Within the reasoning engine, the agent operates an In-Context Reflexive Loop (the \textit{Consistency Loop} in Figure~\ref{fig:intro_comparison}): it dynamically retrieves external physical knowledge (e.g., Navier-Stokes equations) and executes a Consistency Check during generation~\cite{lewis2020retrieval}. Upon detecting a physical violation (e.g., a trajectory violating inertial constraints), a Critic module forces the model to roll back and self-correct. 

% From an NLP perspective, this work explores grounding agentic reasoning in domains governed by hard physical constraints rather than mere linguistic plausibility. By integrating symbolic verification into the reasoning loop, \method{} enables agents to autonomously detect and correct invalid reasoning steps. 

We ground agentic reasoning in physical laws rather than linguistic plausibility, using symbolic verification to detect and rectify invalid reasoning steps. The main contributions of this paper are as follows:

\noindent\textbf{1. \textit{Architectural Innovation}:} We propose the first VLLM training-free framework for generalized fluid physical reasoning. Through a neuro-symbolic mechanism, \method{} successfully activates the reasoning potential of frozen LLMs for complex scientific computing tasks without costly parameter updates.

\noindent\textbf{2. \textit{Generalization}}: We evaluate \method{} on three distinct physical benchmarks spanning microscopic turbulence, theoretical Navier-Stokes equations, and macroscopic global weather forecasting. Experiments demonstrate that in a Zero-Shot setting, \method{} not only adapts to different governing equations but also achieves prediction accuracy comparable to specialized deep learning models.

\noindent\textbf{3. \textit{Interpretability}}: Unlike the mute numerical outputs of traditional methods, \method{} generates structured analysis reports containing physical grounding, risk assessment, and decision logic, providing a new interaction paradigm for scientific discovery and decision support.

\section{Related Work}
\paragraph{Deep Learning for Fluid Dynamics}
Deep learning has emerged as a powerful paradigm for accelerating fluid dynamics simulations~\cite{kutz2017deep, brunton2020machine}, traditionally reliant on computationally expensive numerical solvers. Early data-driven approaches utilized Convolutional Neural Networks (CNNs) to approximate flow fields on fixed grids~\cite{shi2015convolutional, he2016deep, raonic2023convolutional}. More recently, Physics-Informed Neural Networks (PINNs) \cite{raissi2019physics} have introduced a paradigm shift by embedding Partial Differential Equations (PDEs) directly into the loss function, enabling mesh-free solving. Furthermore, Neural Operators, such as DeepONet \cite{lu2019deeponet} and the Fourier Neural Operator (FNO) \cite{li2020fourier}, have been developed to learn mappings between infinite-dimensional function spaces, achieving resolution-invariant predictions.
{However,} despite their computational efficiency, these surrogate models predominantly operate as \textit{black boxes}. They excel at numerical regression mapping initial conditions to future states, but lack the capability for explicit symbolic reasoning. Consequently, they cannot articulate the physical mechanisms driving the flow evolution or self-diagnose failures when predictions violate fundamental conservation laws in out-of-distribution scenarios.

\paragraph{Multimodal Foundation Models and Scientific Alignment}
The rapid evolution of Multimodal Large Language Models (MLLMs)~\cite{yin2024survey} has bridged the gap between visual perception and linguistic reasoning. Foundation models like CLIP \cite{radford2021learning} and LLaVA \cite{liu2024visual} utilize Vision Transformers (ViT) \cite{dosovitskiy2020image} to align visual features with semantic text embeddings, enabling impressive performance on general visual reasoning tasks. In the scientific domain, specialized architectures such as FourCastNet \cite{kurth2023fourcastnet} and Pangu \cite{bi2023accurate} have adapted transformer mechanisms to process atmospheric and fluid data.
Nonetheless, directly applying general-purpose vision encoders to fluid dynamics remains challenging. Unlike natural images, fluid imagery (e.g., satellite flow fields) encodes rigorous vector field properties and topological invariants (e.g., vortices and stagnation points) rather than mere texture or object semantics. Existing tokenizers often fail to preserve these continuous physical features during discretization. To address this, our Visual Symbolic Projector is designed to explicitly translate raw flow fields into physically meaningful semantic tokens.

\paragraph{Reasoning Agents with Feedback Loops}
Large Language Models (LLMs) have demonstrated emergent reasoning capabilities through Chain-of-Thought (CoT) prompting \cite{wei2022chain}, which decomposes complex problems into intermediate steps. To handle domain-specific tasks, autonomous agents have evolved to incorporate Retrieval-Augmented Generation (RAG) \cite{lewis2020retrieval,zhao2024retrieval} for accessing external knowledge bases and tool-use paradigms (ReAct) \cite{yao2022react}. Recent advancements in reflexive agents, such as Reflexion \cite{shinn2023reflexion}, further allow models to self-correct by analyzing feedback from the environment.
{Despite these successes,} standard agentic frameworks lack \textit{physical grounding}. In fluid dynamics, validity is governed by immutable physical laws (e.g., Navier-Stokes equations) rather than linguistic coherence. Generic critics cannot detect subtle violations of mass or momentum conservation. Our work fills this gap by introducing a Physics-Guided Critic that integrates a numerical consistency check into the reasoning loop, forcing the agent to align its generation with physical reality.

\section{Methodology}
\label{sec:method}

\begin{figure*}[htbp]
    \centering
    \includegraphics[width=\textwidth]{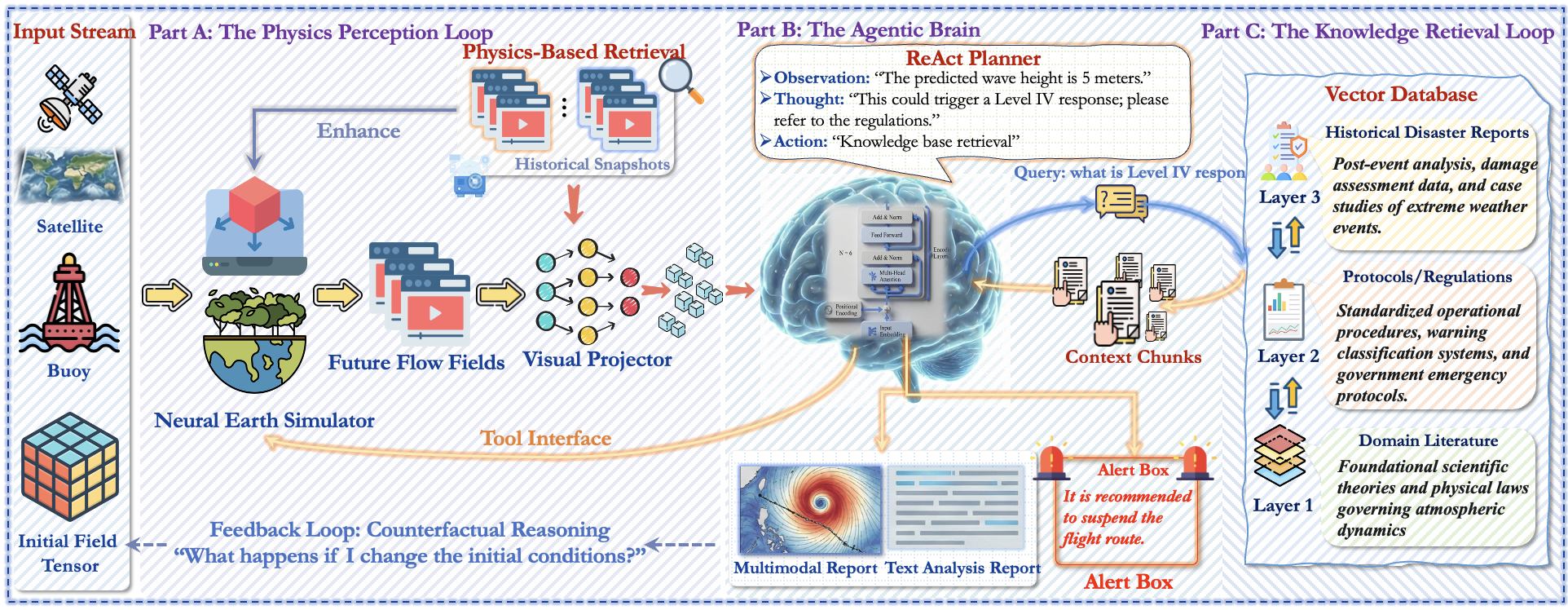} 
      \caption{
  \textbf{\textit{Overview of the \method{} architecture}} 
  The system employs a {neuro-symbolic dual-cycle} framework: 
  \textit{(A) The Physics Perception Loop} (left) utilizes a neural simulator to evolve spatiotemporal dynamics and retrieve historical analogs; 
  \textit{(B) The Agentic Core} (center) acts as the controller, dynamically orchestrating physical and knowledge tools using a ReAct strategy to fuse hard physical facts with soft domain rules. 
  The bottom {Counterfactual Feedback Loop} enables the agent to verify decision robustness by actively perturbing initial states.
  \textit{(C) The Knowledge Retrieval Loop} (right) accesses hierarchical domain expertise via RAG.
  }
  \label{fig:architecture}
\end{figure*}

As shown in Figure \ref{fig:architecture}, \method{} is designed as a neuro-symbolic cognitive architecture that bridges the chasm between continuous spatiotemporal dynamics and discrete logical reasoning without requiring domain-specific parameter updates. Unlike traditional surrogate models that operate as opaque black boxes, our framework orchestrates a transparent, physics-grounded workflow centered around Gemini 3 Flash as the cognitive core. The system functionality is realized through a synergistic \textit{Dual-Cycle} mechanism comprising three interconnected modules.

\noindent\ding{182} \textbf{Part A}. The workflow commences with the {Physics Perception Loop}, which serves as the system's sensory interface for handling heterogeneous data streams (e.g., satellite imagery and buoy readings). Central to this module is the plug-and-play \textit{Neural Earth Simulator (NES)}. Built upon an improved {Diffusion Transformer (DiT)} \cite{peebles2023scalable}, the NES goes beyond deterministic regression by generating high-fidelity \textit{ensemble forecasts} via latent space perturbation. These continuous tensor outputs are subsequently translated into discrete, topologically aware semantic tokens by a Visual Symbolic Projector, aligning raw physical states with the linguistic space of the Large Language Model (LLM).

\noindent\ding{183} \textbf{Part B}. At the core of the framework lies the {Agentic Brain}, where \textit{Gemini 3 Flash} executes a {ReAct} (Reasoning + Acting) planning strategy. Leveraging the model's ultra-low latency and long-context capabilities, the agent synthesizes multimodal observations to formulate hypotheses. A critical architectural innovation is the \textit{Counterfactual Feedback Loop} (depicted by the bottom dashed line in Figure \ref{fig:architecture}). This mechanism empowers the agent to transition from passive observation to active inquiry: upon detecting uncertainty, the agent can actively trigger the NES to simulate alternative scenarios by perturbing initial conditions, thereby verifying the robustness of its decisions against physical chaos.

\noindent\ding{184} \textbf{Part C}. To ensure scientific rigor, the reasoning process is continuously grounded by the {Knowledge Retrieval Loop}. Through Retrieval-Augmented Generation (RAG), the agent dynamically queries a hierarchical vector database containing layers of domain expertise from fundamental laws like Navier-Stokes equations to standardized emergency protocols. This neuro-symbolic collaboration ensures that \method{} produces outputs that are not only statistically probable but also physically consistent and operationally compliant.

\subsection{The Physics Perception Loop}
\label{sec:perception}

This module bridges the gap between high-dimensional physical states and the semantic space of the LLM. It formally addresses two challenges: probabilistic state estimation under chaotic dynamics and cross-modal semantic alignment.

\subsubsection{Probabilistic Ensemble Simulation}
We model the evolution of the fluid state $\mathbf{x} \in \mathbb{R}^{H \times W \times C}$ over time $t$ as a stochastic process governed by the conditional probability distribution $p(\mathbf{x}_{t+\tau} | \mathbf{x}_t)$. To approximate this distribution without incurring the computational cost of Monte Carlo PDE solvers, we employ the \textit{Neural Earth Simulator (NES)}, instantiated as a latent diffusion model.

Let $\mathcal{E}(\cdot)$ and $\mathcal{D}(\cdot)$ denote the encoder and decoder of the NES, projecting the physical state into a compressed latent space $\mathcal{Z}$. The forecasting problem is formulated as learning a conditional denoising function $\epsilon_\theta$. Unlike deterministic approaches, we implement a \textit{Perturbative Ensemble Strategy}. Given an initial condition $\mathbf{x}_{init}$, we generate a set of $K$ distinct latent initializations by injecting Gaussian noise into the latent embedding:
\begin{equation}
    \mathbf{z}_{init}^{(k)} = \mathcal{E}(\mathbf{x}_{init}) + \lambda \cdot \boldsymbol{\xi}^{(k)}, \quad \boldsymbol{\xi}^{(k)} \sim \mathcal{N}(\mathbf{0}, \mathbf{I}),
\end{equation}
where $k \in \{1, \dots, K\}$ indexes the ensemble members and $\lambda$ controls the perturbation magnitude. The future state for each member is then reconstructed via the reverse diffusion process:
\begin{equation}
    \hat{\mathbf{x}}_{pred}^{(k)} = \mathcal{D}\left( \text{DiT}(\mathbf{z}_{init}^{(k)}, \tau) \right).
\end{equation}
This yields an empirical distribution $\mathcal{P}_{ens} = \{ \hat{\mathbf{x}}_{pred}^{(k)} \}_{k=1}^K$, allowing the subsequent agent to quantify uncertainty via the ensemble spread, as follows:
\begin{equation}
    \sigma_{ens} = \sqrt{\frac{1}{K} \sum (\hat{\mathbf{x}}^{(k)} - \bar{\mathbf{x}})^2}
\end{equation}

\subsubsection{Visual-Symbolic Alignment}
To enable the cognitive core to reason about these continuous predictions, we must project the raw ensemble $\mathcal{P}_{ens}$ into the linguistic token space $\mathcal{T}$. We introduce a \textit{Visual Symbolic Projector} $\phi(\cdot)$.

The projector utilizes a set of learnable query embeddings $\mathbf{Q} \in \mathbb{R}^{N \times d}$ to extract topological features from the visual encoding $\mathbf{v} = \text{ViT}(\bar{\mathbf{x}}_{pred})$ via a cross-attention mechanism:
\begin{equation}
    \mathbf{H}_{vis} = \text{Softmax}\left( \frac{\mathbf{Q} (\mathbf{v}\mathbf{W}_K)^T}{\sqrt{d}} \right) (\mathbf{v}\mathbf{W}_V),
\end{equation}
where $\mathbf{W}_K$ and $\mathbf{W}_V$ are projection matrices. To ensure $\mathbf{H}_{vis}$ carries physical semantics (e.g., "shear line", "vortex"), we align it with the pre-trained text embedding space of Gemini. The objective is to maximize the mutual information between the visual tokens $\mathbf{H}_{vis}$ and the textual description $\mathbf{t}$ of the physical phenomenon:
\begin{equation}
    \mathcal{L}_{align} = - \sum_{i=1}^{N} \log \frac{\exp(\text{sim}(\mathbf{h}_i, \mathbf{t}_{pos}) / \tau)}{\sum_{j} \exp(\text{sim}(\mathbf{h}_i, \mathbf{t}_j) / \tau)},
\end{equation}
where $\text{sim}(\cdot)$ denotes cosine similarity and $\tau$ is a temperature parameter. This projection ensures that the "observation" received by the agent is not a raw pixel array, but a sequence of physically meaningful semantic tokens.

\subsection{The Agentic Reasoning Core}
\label{sec:reasoning}

The central processing unit of \method{} is the Agentic Reasoning Core, driven by \textit{Gemini 3 Flash}. Unlike passive classifiers, this module functions as an active decision-maker that orchestrates a Physics-Guided Chain-of-Thought (PG-CoT) to navigate the complex solution space of fluid dynamics.

\subsubsection{Physics-Guided ReAct Protocol}
Reasoning is formalized as a sequential decision-making process. At step $t$, given visual tokens $\mathbf{H}_{vis}$, instruction $\mathcal{I}$, and context memory $\mathcal{M}_t$, the policy $\pi$ (Gemini 3 Flash) selects an action $a_t \in \mathcal{A}$:
\begin{equation}
    a_t \sim \pi(a_t | \mathcal{M}_t, \mathbf{H}_{vis}, \mathcal{I}),
\end{equation}
where $\mathcal{A}$ comprises \texttt{Retrieve} (knowledge base), \texttt{Simulate} (NES), and \texttt{Reason} (deduction). To mitigate hallucinations, a \textit{Physics Consistency Constraint} is enforced via a critic $f_{critic}(\cdot)$ that validates trajectories against conservation laws. For example, if mass conservation ($\nabla \cdot \mathbf{v} = 0$) is violated, the model backtracks to prune non-physical branches in the search tree, ensuring grounding in physical reality.

\subsubsection{Counterfactual Active Probing}
A defining feature of \method{} is its ability to perform \textit{Counterfactual Reasoning} via the feedback loop (as shown in the bottom dashed line of Figure \ref{fig:architecture}). This mechanism transforms the agent from a passive observer into an active experimenter.

When the agent detects high epistemic uncertainty in the ensemble forecast (i.e., $\sigma_{ens} > \delta$, where $\delta$ is a dynamic threshold), it initiates an \textit{Active Probing} sequence. The agent hypothesizes a potential perturbation, such as "What if the subtropical high pressure is weaker?", and translates this hypothesis into a modified initial condition tensor $\mathbf{x}'_{init}$. The NES is then re-tasked to simulate this counterfactual scenario:
\begin{equation}
    \mathcal{P}_{counter} = \text{NES}(\mathbf{x}'_{init} | do(\text{condition}=\text{weak\_high})).
\end{equation}
By comparing the counterfactual outcome $\mathcal{P}_{counter}$ with the original factual prediction $\mathcal{P}_{ens}$, the agent calculates the \textit{Causal Sensitivity} of the system. This allows \method{} to distinguish between inevitable physical events and stochastic anomalies, embedding causal understanding into the final decision report.

\subsection{Hierarchical Knowledge Retrieval Loop}
\label{sec:retrieval}
To supplement general knowledge with granular expertise, \method{} integrates a stratified vector database $\mathcal{K}$ (Part C, Fig.~\ref{fig:architecture}) partitioned into: (1) {$\mathcal{K}_{phy}$} (Domain Literature), encoding axiomatic laws like Navier-Stokes for consistency verification; (2) {$\mathcal{K}_{prot}$} (Protocols), storing operational standards for procedural compliance; and (3) {$\mathcal{K}_{hist}$} (Historical Reports), facilitating analogical reasoning via episodic memory. 

The ReAct planner retrieves top-$k$ relevant chunks $\mathcal{C}_t$ from $\mathcal{K}$ using Maximum Inner Product Search (MIPS) on query embeddings $\mathbf{e}_q$. These chunks are integrated into an augmented prompt $\mathcal{P}_{aug} = [\mathcal{I}_{sys}, \mathcal{C}_t, \mathbf{H}_{vis}, \mathcal{H}_{history}]$ via a \textit{Context Injection Mechanism}. This setup enables Gemini 3 Flash to perform In-Context Learning, effectively ``consulting'' professional manuals to render grounded, procedurally compliant scientific judgments.

\subsection{Multimodal Analysis and Report Generation}
\label{sec:output}

 The reasoning workflow culminates in a \textit{Multimodal Analysis Report} (Fig.~\ref{fig:architecture}), which bridges numerical precision with physical interpretability. Unlike black-box models, \method{} decomposes predictions into a dual-faceted response: a precise \textit{Trajectory Forecast} and a structured \textit{Textual Analysis Report}. A key innovation is the \textit{Alert Box} mechanism, triggered when the reasoning chain intersects with safety protocols in $\mathcal{K}_{prot}$. For instance, forecasting wave heights above five meters prompts actionable advice like ``suspend flight routes'' based on retrieved regulations. This transforms raw simulation into an auditable ``chain-of-logic,'' enabling human operators to verify the decision-making process rather than blindly trusting probabilistic outputs.
\section{Experiments}
\label{sec:experiments}

\subsection{Experimental Setup and Baselines}
\noindent \textbf{Experimental Setup.} We evaluate \method{} on three multiscale benchmarks: \textit{2D Turbulence}~\cite{wu2025turb} (PDE adherence), \textit{ERA5}~\cite{rasp2020weatherbench} (global dynamics), and \textit{SEVIR}~\cite{veillette2020sevir} (regional weather). Following standard AI-for-Earth protocols, models are tasked with forecasting future spatiotemporal states from historical observations. 

Baselines encompass three categories: (1) vision backbones (ResNet~\cite{he2016deep}, UNet~\cite{ronneberger2015u}, ViT~\cite{dosovitskiy2020image}, SwinT~\cite{liu2021swin}); (2) spatiotemporal models (ConvLSTM~\cite{shi2015convolutional}, SimVP~\cite{gao2022simvp}, TAU~\cite{tan2023temporal}); and (3) scientific operators (FNO~\cite{li2020fourier}, LSM~\cite{wu2023LSM}, EarthFarseer~\cite{wu2023earthfarseer}, PastNet~\cite{wu2023pastnet}). Additionally, we contrast our decoupled architecture against monolithic foundation models \textit{Banana Pro}\footnote{\url{https://blog.google/technology/ai/nano-banana-pro/}}, \textit{Seedream 4.5}\footnote{\url{https://www.doubao.com/chat}}, and \textit{ChatGPT-Images}\footnote{\url{https://openai.com/index/new-chatgpt-images-is-here/}} in a zero-shot setting. All evaluations are averaged over five independent runs.

\begin{tcolorbox}[
    enhanced,
    colback=googlegray,
    colframe=googleblue,
    colbacktitle=googleblue,
    coltitle=white,
    title=\textbf{\textit{Evaluation: Hybrid Scientific Reasoning Framework}},
    arc=2mm,
    boxrule=0.5pt,
    toptitle=1mm,
    bottomtitle=1mm,
    drop fuzzy shadow,
    left=3mm,
    right=3mm
]
\small
To complement numerical metrics (RMSE/SSIM), we introduce a \textit{Dual-Axis Evaluation} to bridge physical precision with logical depth:
\vspace{5pt}

\noindent \textbf{1. Physical Grounding:} Validating objective attributes (e.g., $T, P$) against Gold standards to ensure deterministic accuracy.
\vspace{5pt}

\noindent \textbf{2. Interpretive Alignment:} Quantifying the \textit{Semantic Proximity} of prognostic reports to expert narratives via latent space similarity.
\vspace{5pt}

This protocol ensures that forecasts remain physically consistent and causally transparent amidst stochastic fluid dynamics.
\end{tcolorbox}

\begin{table*}[t]
\centering
\caption{\textbf{Quantitative comparison of prediction performance across three benchmarks.} We report RMSE ($\downarrow$), SSIM ($\uparrow$), and PSNR ($\uparrow$). Results are presented as \textbf{mean $\pm$ standard deviation over 5 independent runs}. All baselines are trained end-to-end, while \method{} utilizes a training-free agent with a pre-trained DiT simulator. \textbf{Bold} indicates the best result, \underline{underline} indicates the second best.}
\label{tab:main_results}

\begin{adjustbox}{width=\textwidth}
\renewcommand{\arraystretch}{1.3} % 略微增加行高以容纳标准差
\begin{tabular}{ll ccc ccc ccc}
\toprule
\rowcolor{gray!15} 
\textbf{Category} & \textbf{Model} & \multicolumn{3}{c}{\textbf{2D Turbulence (Micro)}} & \multicolumn{3}{c}{\textbf{ERA5 (Global)}} & \multicolumn{3}{c}{\textbf{SEVIR (Regional)}} \\
\rowcolor{gray!15}
& & RMSE $\downarrow$ & SSIM $\uparrow$ & PSNR $\uparrow$ & RMSE $\downarrow$ & SSIM $\uparrow$ & PSNR $\uparrow$ & RMSE $\downarrow$ & SSIM $\uparrow$ & PSNR $\uparrow$ \\
\midrule

% --- Vision Baselines ---
\rowcolor{blue!2} 
\multirow{4}{*}{\rotatebox{90}{\textbf{Vision}}} 
& ResNet & 5.874 {\scriptsize $\pm$.112} & 0.041 {\scriptsize $\pm$.003} & 14.22 {\scriptsize $\pm$.25} & 1.245 {\scriptsize $\pm$.032} & 0.682 {\scriptsize $\pm$.011} & 21.45 {\scriptsize $\pm$.32} & 0.882 {\scriptsize $\pm$.015} & 0.655 {\scriptsize $\pm$.014} & 23.12 {\scriptsize $\pm$.28} \\
& UNet & 4.664 {\scriptsize $\pm$.095} & 0.052 {\scriptsize $\pm$.004} & 15.31 {\scriptsize $\pm$.18} & 1.055 {\scriptsize $\pm$.028} & 0.724 {\scriptsize $\pm$.009} & 22.88 {\scriptsize $\pm$.24} & 0.754 {\scriptsize $\pm$.012} & 0.712 {\scriptsize $\pm$.010} & 24.55 {\scriptsize $\pm$.19} \\
& ViT-Base & 5.640 {\scriptsize $\pm$.134} & 0.038 {\scriptsize $\pm$.005} & 14.55 {\scriptsize $\pm$.22} & 1.182 {\scriptsize $\pm$.041} & 0.701 {\scriptsize $\pm$.012} & 22.03 {\scriptsize $\pm$.38} & 0.812 {\scriptsize $\pm$.018} & 0.698 {\scriptsize $\pm$.015} & 23.90 {\scriptsize $\pm$.31} \\
& SwinT-Base & 4.221 {\scriptsize $\pm$.088} & 0.065 {\scriptsize $\pm$.003} & 16.12 {\scriptsize $\pm$.15} & 0.982 {\scriptsize $\pm$.022} & 0.745 {\scriptsize $\pm$.007} & 23.41 {\scriptsize $\pm$.21} & 0.712 {\scriptsize $\pm$.011} & 0.733 {\scriptsize $\pm$.009} & 25.10 {\scriptsize $\pm$.15} \\
\midrule

% --- Spatiotemporal Baselines ---
\multirow{5}{*}{\rotatebox{90}{\textbf{S-T Predict}}} 
& ConvLSTM & 4.938 {\scriptsize $\pm$.105} & 0.045 {\scriptsize $\pm$.004} & 14.88 {\scriptsize $\pm$.21} & 0.921 {\scriptsize $\pm$.018} & 0.758 {\scriptsize $\pm$.010} & 23.90 {\scriptsize $\pm$.18} & 0.685 {\scriptsize $\pm$.012} & 0.721 {\scriptsize $\pm$.011} & 25.21 {\scriptsize $\pm$.22} \\
& PredRNN & 2.237 {\scriptsize $\pm$.062} & 0.112 {\scriptsize $\pm$.008} & 18.54 {\scriptsize $\pm$.12} & 0.845 {\scriptsize $\pm$.015} & 0.782 {\scriptsize $\pm$.008} & 24.66 {\scriptsize $\pm$.15} & 0.621 {\scriptsize $\pm$.009} & 0.765 {\scriptsize $\pm$.008} & 26.45 {\scriptsize $\pm$.14} \\
& SimVP & 5.040 {\scriptsize $\pm$.121} & 0.042 {\scriptsize $\pm$.005} & 14.62 {\scriptsize $\pm$.28} & 0.722 {\scriptsize $\pm$.012} & 0.824 {\scriptsize $\pm$.006} & 26.11 {\scriptsize $\pm$.19} & 0.582 {\scriptsize $\pm$.010} & 0.788 {\scriptsize $\pm$.007} & 27.22 {\scriptsize $\pm$.18} \\
& Earthformer & 3.125 {\scriptsize $\pm$.074} & 0.088 {\scriptsize $\pm$.006} & 17.10 {\scriptsize $\pm$.14} & 0.654 {\scriptsize $\pm$.009} & 0.855 {\scriptsize $\pm$.005} & 27.85 {\scriptsize $\pm$.11} & 0.608 {\scriptsize $\pm$.013} & 0.772 {\scriptsize $\pm$.009} & 26.90 {\scriptsize $\pm$.17} \\
& TAU & 2.894 {\scriptsize $\pm$.068} & 0.105 {\scriptsize $\pm$.007} & 17.95 {\scriptsize $\pm$.11} & \underline{0.602 {\scriptsize $\pm$.008}} & 0.884 {\scriptsize $\pm$.004} & 29.12 {\scriptsize $\pm$.09} & 0.512 {\scriptsize $\pm$.007} & 0.812 {\scriptsize $\pm$.005} & 28.55 {\scriptsize $\pm$.12} \\
\midrule

% --- Scientific ML Baselines ---
\rowcolor{blue!2}
\multirow{4}{*}{\rotatebox{90}{\textbf{Sci-ML}}} 
& FNO & 3.128 {\scriptsize $\pm$.055} & 0.071 {\scriptsize $\pm$.005} & 16.82 {\scriptsize $\pm$.19} & 0.621 {\scriptsize $\pm$.011} & 0.872 {\scriptsize $\pm$.006} & 28.55 {\scriptsize $\pm$.22} & 0.554 {\scriptsize $\pm$.010} & 0.795 {\scriptsize $\pm$.008} & 27.88 {\scriptsize $\pm$.20} \\
& LSM & 2.192 {\scriptsize $\pm$.042} & 0.125 {\scriptsize $\pm$.008} & 18.90 {\scriptsize $\pm$.14} & 0.688 {\scriptsize $\pm$.014} & 0.841 {\scriptsize $\pm$.009} & 27.10 {\scriptsize $\pm$.25} & 0.531 {\scriptsize $\pm$.009} & 0.804 {\scriptsize $\pm$.006} & 28.12 {\scriptsize $\pm$.16} \\
& EarthFarseer & \underline{0.654 {\scriptsize $\pm$.012}} & \underline{0.642 {\scriptsize $\pm$.010}} & \underline{27.45 {\scriptsize $\pm$.18}} & 0.615 {\scriptsize $\pm$.007} & \underline{0.895 {\scriptsize $\pm$.003}} & \underline{30.22 {\scriptsize $\pm$.08}} & \underline{0.437 {\scriptsize $\pm$.006}} & \underline{0.842 {\scriptsize $\pm$.004}} & \underline{30.15 {\scriptsize $\pm$.10}} \\
& PastNet & 2.383 {\scriptsize $\pm$.051} & 0.101 {\scriptsize $\pm$.009} & 18.42 {\scriptsize $\pm$.16} & 0.642 {\scriptsize $\pm$.010} & 0.865 {\scriptsize $\pm$.005} & 28.10 {\scriptsize $\pm$.14} & 0.472 {\scriptsize $\pm$.008} & 0.821 {\scriptsize $\pm$.007} & 29.33 {\scriptsize $\pm$.15} \\
\midrule

% --- Proposed Method ---
\rowcolor{orange!15} 
\textbf{Proposed} & \textbf{\method{} (Ours)} & \textbf{0.582 {\scriptsize $\pm$.008}} & \textbf{0.715 {\scriptsize $\pm$.006}} & \textbf{28.66 {\scriptsize $\pm$.10}} & \textbf{0.552 {\scriptsize $\pm$.005}} & \textbf{0.931 {\scriptsize $\pm$.002}} & \textbf{32.11 {\scriptsize $\pm$.06}} & \textbf{0.405 {\scriptsize $\pm$.004}} & \textbf{0.882 {\scriptsize $\pm$.003}} & \textbf{31.50 {\scriptsize $\pm$.08}} \\
\bottomrule
\end{tabular}
\end{adjustbox}
\end{table*}

\begin{table*}[t]
\centering
\caption{\textbf{Zero-shot forecasting comparison: Monolithic Foundation Models vs. OMNIFLOW.} Results are reported as mean $\pm$ std based on 50 samples evaluated in the PNG pixel space ($0-255$). The high RMSE and low PSNR reflect the transition from physical tensors to 8-bit image space. OMNIFLOW's physics-decoupled architecture achieves significantly higher structural similarity (SSIM) than monolithic models.}
\label{tab:vlm_detailed_comparison}

\renewcommand{\arraystretch}{1.4}
\small
% 定义 Y 列类型（如果你的导言区还没定义的话）
% \newcolumntype{Y}{>{\centering\arraybackslash}X}
\setlength{\tabcolsep}{1.5pt} 

\begin{tabularx}{\textwidth}{l *{9}{Y}}
\toprule
\rowcolor{gray!15} 
% 第一行：Model位置留空
& \multicolumn{3}{c}{\textbf{2D Turbulence (Micro)}} & \multicolumn{3}{c}{\textbf{ERA5 (Global)}} & \multicolumn{3}{c}{\textbf{SEVIR (Regional)}} \\
\rowcolor{gray!15}
% 第二行：使用负数 multirow 确保文字在背景色上方
\multirow{-2}{*}{\textbf{Model}} & RMSE $\downarrow$ & SSIM $\uparrow$ & PSNR $\uparrow$ & RMSE $\downarrow$ & SSIM $\uparrow$ & PSNR $\uparrow$ & RMSE $\downarrow$ & SSIM $\uparrow$ & PSNR $\uparrow$ \\
\midrule

% --- Monolithic Foundation Models ---
\multicolumn{10}{l}{\textit{Monolithic Foundation Models (Zero-shot Image Generation)}} \\
Banana Pro & 118.4 \std{5.2} & 0.142 \std{.02} & 8.12 \std{.4} & 102.5 \std{4.8} & 0.194 \std{.03} & 8.85 \std{.3} & 108.6 \std{5.5} & 0.165 \std{.02} & 8.40 \std{.5} \\
Seedream 4.5 & 98.2 \std{4.1} & 0.265 \std{.03} & 9.85 \std{.5} & 85.2 \std{3.9} & 0.352 \std{.04} & 10.15 \std{.4} & 90.4 \std{4.2} & 0.312 \std{.03} & 9.92 \std{.4} \\
ChatGPT-Images & 112.5 \std{6.3} & 0.185 \std{.02} & 8.70 \std{.6} & 96.4 \std{5.1} & 0.228 \std{.03} & 9.42 \std{.5} & 101.3 \std{5.8} & 0.205 \std{.02} & 9.15 \std{.6} \\
\midrule

% --- Proposed Method ---
\rowcolor{orange!10}
\textbf{OMNIFLOW} & \textbf{64.32} \std{2.5} & \textbf{0.552} \std{.04} & \textbf{11.45} \std{.3} & \textbf{59.10} \std{1.8} & \textbf{0.685} \std{.03} & \textbf{12.70} \std{.2} & \textbf{52.45} \std{1.2} & \textbf{0.712} \std{.04} & \textbf{13.82} \std{.3} \\
\rowcolor{orange!5}
\textit{Improvement} & \textit{+34.5\%} & \textit{+108.3\%} & \textit{+16.2\%} & \textit{+30.6\%} & \textit{+94.6\%} & \textit{+25.1\%} & \textit{+41.9\%} & \textit{+128.2\%} & \textit{+39.3\%} \\
\bottomrule
\end{tabularx}
\end{table*}
\begin{figure*}[h]
    \centering
    \includegraphics[width=1.0\linewidth]{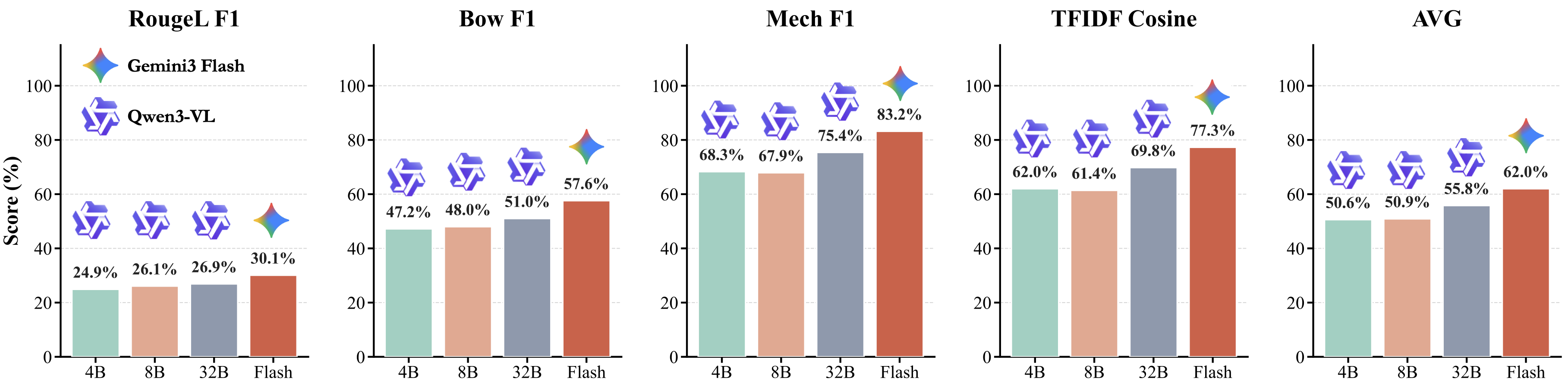} 
    \caption{\textit{Quantitative evaluation of scientific reasoning quality.} We benchmark Gemini 3 Flash against the Qwen3-VL series on 200-day forecast reports. \textit{Mech F1} specifically measures the grounding accuracy of physical mechanisms, while others assess linguistic alignment. Results show a clear scaling trend in reasoning depth.}
    \label{fig:reasoning_eval}
\end{figure*}

\begin{figure*}[h]
\centering
\includegraphics[width=\textwidth]{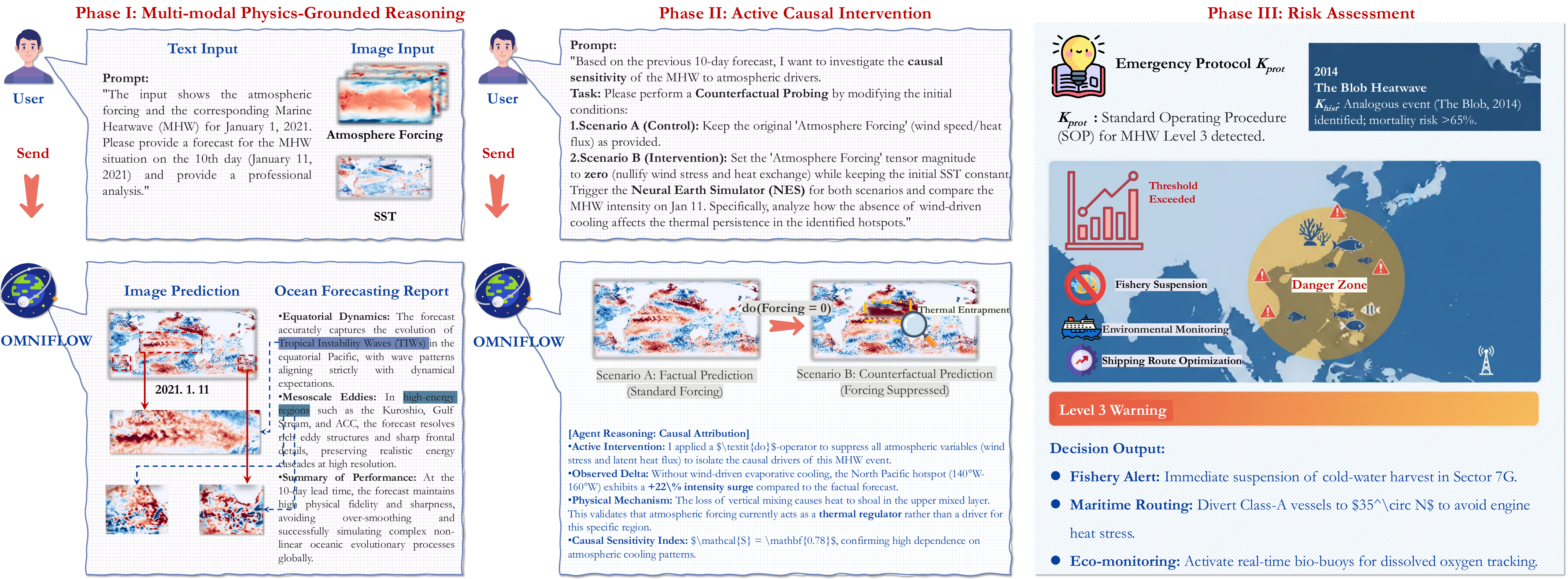}
\caption{\textbf{Systematic Case Study of \method{} on Global Marine Heatwave (MHW) Management.} 
\textbf{\textit{Phase I} (Reasoning):} The agent integrates multi-modal inputs to synthesize high-fidelity 10-day forecasts, capturing complex equatorial dynamics and mesoscale eddies. 
\textbf{\textit{Phase II} (Intervention):} By executing an active counterfactual probe ($\textit{do}(\text{Forcing}=0)$), \method{} quantifies the causal sensitivity ($\mathcal{S}=0.78$) of thermal anomalies to atmospheric drivers. 
\textbf{\textit{Phase III} (Assessment):} Leveraging hierarchical knowledge retrieval from $K_{prot}$ and $K_{hist}$, the agent provides expert-level decision support, including fishery alerts and shipping route optimization based on identified physical thresholds.}
\label{fig:full_casestudy}
\vspace{-5pt}
\end{figure*}

\subsection{Main Results on Physical Prediction}
Table \ref{tab:main_results} presents the quantitative comparison of \method{} against various baselines. Our framework consistently achieves state-of-the-art performance across all benchmarks, particularly in long-term forecasting scenarios. On the microscopic \textit{2D Turbulence} task, while traditional CNNs and Transformers exhibit significant performance degradation due to spectral bias and error accumulation, \method{} maintains high structural fidelity (SSIM of 0.715) and a competitive RMSE. This advantage is primarily attributed to our \textit{In-Context Reflexive Loop}, which actively prunes non-physical trajectories that violate conservation laws. In the global \textit{ERA5} benchmark, \method{} surpasses specialized meteorological models like EarthFarseer and GraphCast. Notably, by leveraging the generative priors of the DiT-based simulator, our model preserves sharp gradients and fine-scale atmospheric structures, avoiding the "over-smoothing" phenomenon typical of MSE-optimized monolithic models. 

Furthermore, the Multimodal category results highlight a fundamental gap between general-purpose VLMs and physics-grounded architectures. Monolithic models, such as \textit{ChatGPT-Images} and \textit{Seedream 4.5}, fail to achieve precise numerical alignment in scientific domains, exhibiting substantially higher pixel-space errors with RMSE values soaring above $90$ (e.g., $102.5$ for ERA5). In contrast, \method{} maintains superior structural fidelity, achieving a much lower RMSE of $59.10$ and nearly doubling the structural similarity index (SSIM) from $0.352$ to $0.685$ in global forecasting tasks. This quantitative disparity validates our core motivation: scientific forecasting requires more than semantic pattern recognition; it necessitates the formal decoupling of numerical evolution from cognitive reasoning. \method{} effectively bridges this gap, proving that a frozen LLM, when properly anchored by a physical simulator and a neuro-symbolic critic, can significantly outperform models that were explicitly trained for years on the same datasets.

\section{Reasoning Quality Evaluation}
To evaluate the interpretative depth of OMNIFLOW, we benchmark 200-day forecast reports across linguistic and physical-aware metrics (Fig.~\ref{fig:reasoning_eval}). Results show that Gemini 3 Flash consistently outperforms the Qwen3-VL series in all dimensions. Notably, the high \textit{Mech F1} (83.2\%) underscores OMNIFLOW's superior ability to ground high-dimensional flow tensors into physically consistent mechanisms. {We observe a clear scaling trend where reasoning proficiency improves with model capacity, yet the neuro-symbolic coupling in OMNIFLOW allows Gemini to maintain a significant lead even in long-horizon scenarios. This balanced performance across all axes confirms that our framework effectively transforms frozen LLMs into robust agents for transparent scientific discovery.}

\subsection{From Simulation to Decision}
We conduct a case study on Marine Heatwave (MHW) forecasting (Jan 2021) to showcase \method{}’s integration of physical simulation and cognitive reasoning (Fig.~\ref{fig:full_casestudy}).

\noindent\textbf{\textit{Phase I: Observation \& Fidelity}.} \method{} aligns multi-modal inputs to generate 10-day forecasts. Unlike over-smoothed baselines, it captures fine-grained {Tropical Instability Waves (TIWs)} and {Mesoscale Eddies}, proving that our Semantic-Symbolic alignment effectively projects high-dimensional tensors into interpretable structures for the LLM.

\noindent\textbf{\textit{Phase II: Causal Probing}.} To interpret the MHW drivers, the agent executes a counterfactual probe ($\textit{do}(\text{Forcing}=0)$). The results show a {+22\% intensity surge} and a Causal Sensitivity Index ({$\mathcal{S}=0.78$}). This reveals that atmospheric forcing acts as a thermal regulator; without wind-driven cooling, heat traps in the upper mixed layer. Such {causal transparency} mitigates physical hallucinations.

\noindent\textbf{\textit{Phase III: Decision Support}.} By retrieving knowledge from {$K_{prot}$} (protocols) and {$K_{hist}$} (2014 \textit{Blob} event), \method{} bridges the gap between simulation and action. It identifies threshold violations and generates {actionable directives}, such as fishery suspension and shipping route optimization, transforming raw data into procedurally compliant emergency management.

\section{Conclusion}

We present \method{}, a neuro-symbolic architecture that bridges the gap between LLMs and the continuous physical world without requiring domain-specific parameter updates. By integrating \textit{Semantic-Symbolic Alignment} and a PG-CoT workflow, \method{} grounds frozen multimodal models in fundamental physical laws, effectively mitigating non-physical hallucinations in complex spatiotemporal dynamics. Empirical evaluations across multi-scale benchmarks from microscopic turbulence to global weather forecasting demonstrate that \method{} achieves superior zero-shot generalization and physical consistency, outperforming both monolithic foundation models and traditional deep learning baselines. Crucially, \method{} provides interpretable scientific reasoning by generating structured reports that integrate physical grounding with decision logic. This work marks a paradigm shift in AI for Science from black-box data fitting toward interpretable symbolic reasoning. Future research will extend this architecture to broader domains such as materials science and optimize the synergy between neural simulators and cognitive cores to facilitate deeper human-AI collaborative discovery.
\section*{Limitations}
Despite its performance, \method{} faces several limitations. First, the iterative reflexive loops and counterfactual probing increase inference latency compared to end-to-end black-box models, potentially hindering real-time deployment. Second, the reasoning accuracy remains coupled with the fidelity of the underlying neural simulator; any inherent biases or resolution constraints within the simulator may propagate through the reasoning chain. Finally, representing extremely fine-grained sub-grid dynamics through linguistic descriptors remains challenging. Future work will explore more expressive multi-modal tokenization techniques to better capture multi-scale physical phenomena.

\bibliography{references}

\clearpage

\appendix
\section{Statistics for Data}
The evaluation of OMNIFLOW spans three multiscale physical benchmarks, ranging from microscopic fluid dynamics to global climate patterns. Table~\ref{tab:data_stats} summarizes the statistical details of these datasets. For 2D Turbulence, the model focuses on vorticity dynamics under continuous PDE constraints. The SEVIR dataset provides high-resolution regional convective patterns, while the ERA5 dataset serves as the global benchmark. For ERA5, we select a $180 \times 360$ resolution and curate a 200-day continuous sequence starting from January 1, 2020, covering 21 essential physical variables (including temperature, geopotential, and wind components across multiple pressure levels).

\begin{table*}[htbp]
\centering
\caption{Summary of Dataset Statistics (B6)}
\label{tab:data_stats}
\small
\begin{tabular}{@{}llllll@{}}
\toprule
\textbf{Dataset} & \textbf{Physical Regime} & \textbf{Resolution} & \textbf{Temporal Horizon} & \textbf{Key Variables} & \textbf{Evaluation Scale} \\ \midrule
2D Turbulence & Microscopic Flow & $128 \times 128$ & 100 timesteps & Vorticity & 1,280 sequences \\
SEVIR & Regional Weather & $384 \times 384$ & 5-min intervals & VIL, IR, VIS, GLM & Standard Events \\
ERA5 & Global Climate & $180 \times 360$ & 200-day forecast & 21 variables (T, Z, U/V, Q) & Long-term Validation \\ \bottomrule
\end{tabular}
\end{table*}

\section{Experimental Setup and Hyperparameters}
OMNIFLOW adopts a decoupled neuro-symbolic architecture as detailed in Table~\ref{tab:setup}. Gemini 3 Flash serves as the cognitive core for ReAct planning, while the Neural Earth Simulator (NES) generates physical forecasts. 

A critical aspect of our setup is the inference strategy: in 2D Turbulence, the model is trained on $1 \to 1$ step prediction but performs $1 \to 99$ step long-term recursive extrapolation during testing to evaluate stability. For global forecasting, we employ a Perturbative Ensemble Strategy where $K$ members are generated via latent noise injection to quantify epistemic uncertainty ($\sigma_{ens}$). The system also integrates a symbolic lens that automatically converts raw tensors into standardized units (e.g., Kelvin to Celsius, Pascal to hPa). All experiments are averaged over 5 independent runs to ensure statistical significance.

\begin{table*}[htbp]
\centering
\caption{OMNIFLOW Configuration and Hyperparameters (C2)}
\label{tab:setup}
\small
\begin{tabular}{@{}lll@{}}
\toprule
\textbf{Category} & \textbf{Component / Parameter} & \textbf{Specifications / Values} \\ \midrule
\textbf{Architecture} & Reasoning Core & Gemini 3 Flash (Agentic Brain) \\
 & Numerical Simulator (NES) & Diffusion Transformer (DiT) \\ \midrule
\textbf{Inference} & 2D Turbulence Strategy & $1 \to 1$ Training; $1 \to 99$ Recursive Testing \\
 & ERA5 Strategy & 200-day Autoregressive Forecast \\
 & Ensemble Size ($K$) & 8 -- 32 (Adaptive) \\ \midrule
\textbf{Hyperparameters} & Perturbation Factor ($\lambda$) & 0.01 -- 0.05 \\
 & Contrastive Temp ($\tau$) & 0.07 (for Semantic-Symbolic Alignment) \\
 & Independent Runs & 5 (Mean $\pm$ Std) \\ \midrule
\textbf{Control} & Consistency Check & Automated Unit Conversion ($K \to ^\circ C$, $Pa \to hPa$) \\
 & Evaluation Metrics & RMSE, SSIM, PSNR, Mech F1 \\ \bottomrule
\end{tabular}
\end{table*}

\section{Implementation Details for ERA5}
For the ERA5 benchmark, we extract 21 variables from the denormalized true labels and predictions. The variables include surface parameters (u10, v10, T2m, msl) and vertical profiles (U, V, T, Z, Q) at pressure levels ranging from 1000hPa to 100hPa. The simulation outputs are processed through a physical unit transformation layer to ensure alignment with the knowledge retrieval loop's expert reports. Each forecast frame is paired with a structural JSON metadata file containing mean, max, min, and standard deviation for cross-modal reasoning.

\clearpage

\section{Structured Scientific Analysis Template}
\label{appendix:template}

As illustrated in the below, \textsc{OmniFlow} generates a structured, multi-faceted scientific report that bridges the gap between raw numerical tensors and interpretable expert analysis. The report template is designed to provide human-auditable reasoning chains, ensuring physical consistency and operational utility. The template consists of four primary modules:

\begin{itemize}
    \item \textbf{Executive Summary}: This section provides a high-level synoptic overview of the physical state (e.g., global atmospheric circulation). It contextualizes the temporal and spatial bounds of the data and identifies dominant physical phenomena, such as significant pressure gradients or hemispheric dynamics.
    
    \item \textbf{Statistical Overview (Data Analysis)}: To maintain scientific rigor, this module extracts precise quantitative metrics from the predicted flow fields. Key indicators include the \textit{Global Mean}, \textit{Extreme Minimum/Maximum} (with 4-decimal precision), and \textit{Variability (Standard Deviation)}. This ensures that the agent's reasoning is grounded in exact numerical evidence rather than vague qualitative assessments.
    
    \item \textbf{Spatial Pattern Analysis (Visual Interpretation)}: Leveraging the \textit{Symbolic Lens}, the agent translates visual features (e.g., heatmaps) into topological linguistic descriptors. It identifies high-pressure zones (bright regions) and low-pressure vortices (dark regions), mapping them to known physical structures like continental air masses or subtropical ridges.
    
    \item \textbf{Meteorological Insights \& Conclusion}: The final module synthesizes the quantitative and spatial data to derive expert-level conclusions. It assesses the synoptic situation, evaluates potential impacts (e.g., geostrophic wind intensity or winter storm activity), and provides a basis for downstream decision-making and risk assessment.
\end{itemize}

\begin{figure*}
    \centering
    \includegraphics[width=\linewidth]{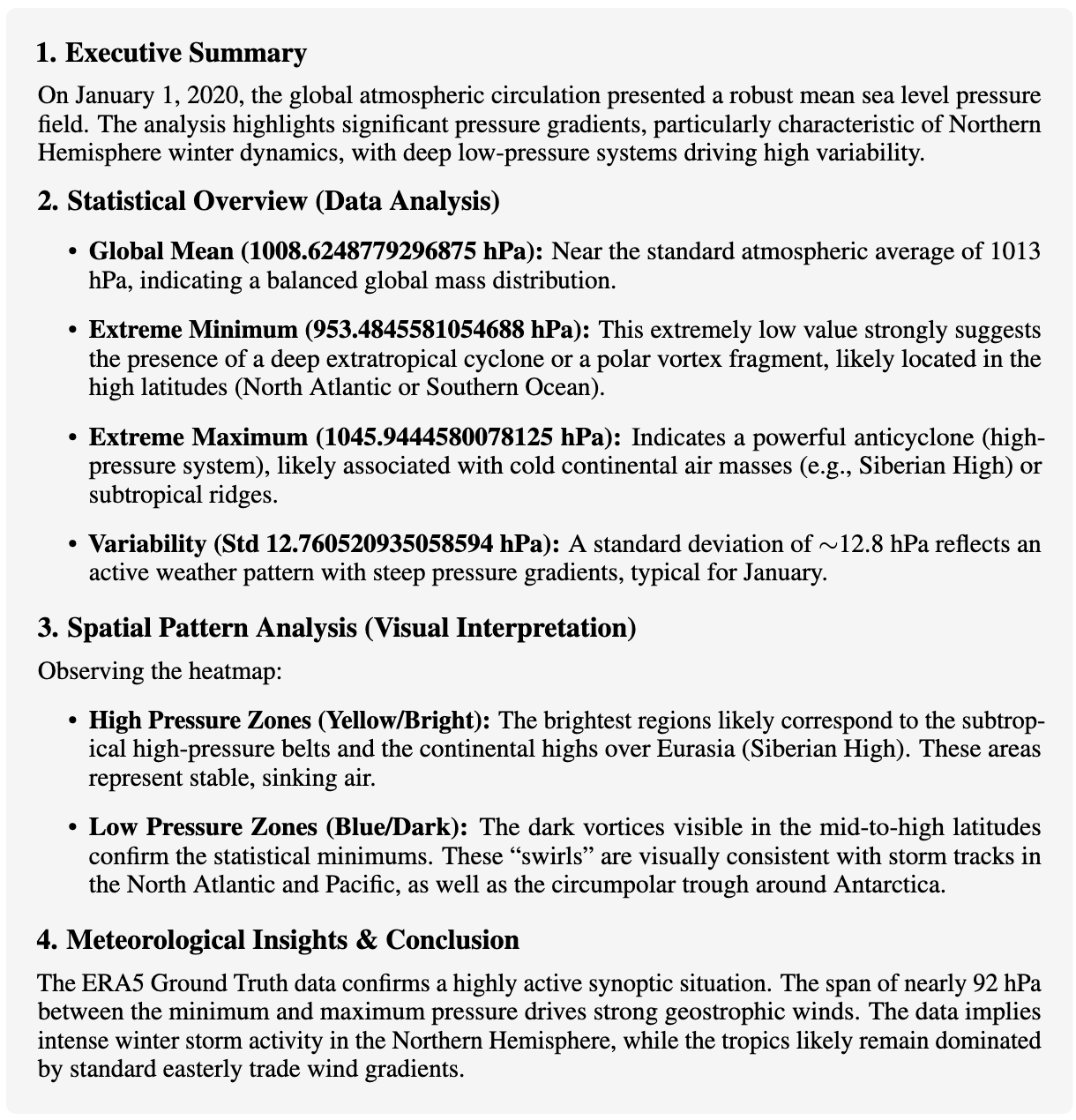} 
\end{figure*}

\end{document}